\documentclass{svproc}
\usepackage{multirow}
\usepackage{array}
\usepackage{graphics}
\usepackage{adjustbox}
\usepackage{cite}
\usepackage{amsmath,amssymb,amsfonts}
\usepackage{algorithmic}
\usepackage{graphicx}
\usepackage{textcomp}
\usepackage{xcolor}
\usepackage{url}

\begin{document}
\mainmatter              
\title{On Significance of Subword tokenization for Low Resource and Efficient Named Entity Recognition: A case study in Marathi}
\titlerunning{Low Resource and Efficient Named Entity Recognition}  
%
\author{Harsh Chaudhari\inst{1,3} \and Anuja Patil\inst{1,3} \and
Dhanashree Lavekar\inst{1,3} \and Pranav Khairnar\inst{1,3} \and Raviraj Joshi\inst{*,2,3} \and Sachin Pande\inst{1} }
\authorrunning{Harsh Chaudhari et al.} 
%
\institute{Pune Institute of Computer Technology, Pune, Maharashtra, India \and
Indian Institute of Technology Madras, Chennai, Tamilnadu, India\and
L3Cube Labs, Pune, Maharashtra, India
\email{\{harshc640,anujadp4,dclavekar,pranavkhairnar016\}@gmail.com, }\email{ravirajoshi@gmail.com}, \email{sspande@pict.edu}
}

\maketitle              

\begin{abstract}
Named Entity Recognition (NER) systems play a vital role in NLP applications such as machine translation, summarization, and question-answering. These systems identify named entities, which encompass real-world concepts like locations, persons, and organizations. Despite extensive research on NER systems for the English language, they have not received adequate attention in the context of low resource languages. In this work, we focus on NER for low-resource language and present our case study in the context of the Indian language Marathi. The advancement of NLP research revolves around the utilization of pre-trained transformer models such as BERT for the development of NER models. However, we focus on improving the performance of shallow models based on CNN, and LSTM by combining the best of both worlds. In the era of transformers, these traditional deep learning models are still relevant because of their high computational efficiency. We propose a hybrid approach for efficient NER by integrating a BERT-based subword tokenizer into vanilla CNN/LSTM models. We show that this simple approach of replacing a traditional word-based tokenizer with a BERT-tokenizer brings the accuracy of vanilla single-layer models closer to that of deep pre-trained models like BERT. We show the importance of using sub-word tokenization for NER and present our study toward building efficient NLP systems. The evaluation is performed on L3Cube-MahaNER dataset using tokenizers from MahaBERT, MahaGPT, IndicBERT, and mBERT. 
\keywords{ Low resource languages, Named Entity Recognition, Deep Learning, Natural Language Processing, Convolutional Neural Network, Bidirectional long short-term memory, Long short-term memory, Marathi NER, Efficient NLP}
\end{abstract}
\section{Introduction}

Named Entity Recognition (NER) plays an very important role in natural language processing (NLP) by identifying and classifying named entities in text \cite{b2,DBLP:journals/corr/ShenYLKA17}. While NER has been extensively studied in various languages, there has been limited attention given to the significance of sub-word tokenization for NER in low-resource languages \cite{b10}. Marathi, as one of the prominent regional languages, holds great importance due to its status as a mother tongue for millions of native speakers \cite{b17}. Furthermore, Marathi is a morphologically rich language, making it challenging to build accurate NER systems. The existing state-of-the-art models often come with high computational costs, rendering them less practical for real-world applications.

The motivation behind this research paper is to bridge the gap between deeper transformer models with high computation costs and shallow deep learning models that cannot capture the intricacies and nuances of the Marathi language. The deeper pre-trained models provide state-of-the-art performance while non-pre-trained shallow deep learning models are computationally efficient but do not achieve top performance. We aim to address this gap by combining the best of both worlds to enhance the performance of shallow models based on CNN (Convolutional Neural Networks), LSTM (Long Short-term Memory Networks), and Bi-LSTM. Through modifications in the tokenization process, we have achieved significant improvements in accuracy, making the performance of these shallow CNN/LSTM models comparable to that of state-of-the-art BERT models.

Subword tokenization has emerged as a promising technique in NLP, splitting words into subword units to capture the morphological structure of a language more effectively \cite{kudo2018subword,kudo2018sentencepiece,sennrich2016neural,10.1007/978-981-16-3690-5_73}. In this paper, we highlight the importance of subword tokenization for the NER task in low resource Marathi language. Moreover, the concept can be extended to other low-resource, morphologically rich languages. Our research presents generic approaches to facilitate the development of more accurate and efficient NER systems for low-resource languages.

\begin{figure}[tbp]
\centerline{\includegraphics[scale=0.5]{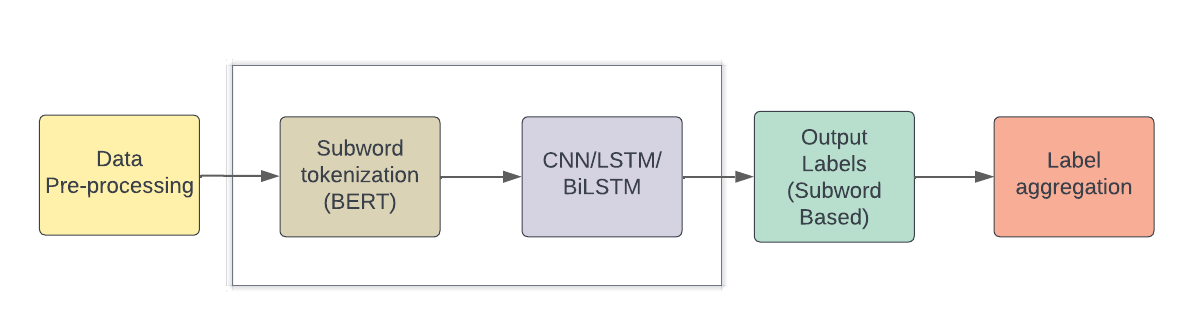}}
\caption{Our Hybrid Approach}
\label{fig:hybrid_approach}
\end{figure}

Our solution aims to tackle the problem of the low resource named entity recognition (NER) by proposing an extremely shallow model with just one layer, achieving high efficiency without compromising performance. We propose a hybrid solution to integrate subword tokenization from BERT into our shallow model as depicted in Figure \ref{fig:hybrid_approach}. This integration allows us to leverage the linguistic coverage of BERT's sub-tokenizer while maintaining the efficiency of our shallow architecture. Additionally, we specifically focus on the Marathi language, which is known for its rich morphology and is susceptible to out-of-vocabulary (OOV) tokens. We make use of subword tokenizers from monolingual Marathi transformer models like L3Cube-MahaBERT\footnote{\url{https://huggingface.co/l3cube-pune/marathi-bert-v2}} \cite{b17}, MahaGPT\footnote{\url{https://huggingface.co/l3cube-pune/marathi-gpt}}, and multilingual BERT models like mBERT and Indic BERT. We show that MahaBERT tokenizer + vanilla CNN works the best on the L3Cube-MahaNER \cite{b10} dataset.

The main contributions of this work are as follows:
We present a hybrid approach for low-resource NER by combining a vanilla single-layer CNN / LSTM model with BERT based sub-word tokenizer.
We present a comparative analysis of different monolingual and multilingual tokenizers for the Marathi language and show that the MahaBERT tokenizer + CNN model works the best.
\section{Related work}

In the early 1990s, the term "Named Entity Recognition" was introduced in the field of Natural Language Processing (NLP) \cite{b11}. Over time, various statistical, machine learning, and deep learning techniques have been developed for NER \cite{b2}.

Regarding specific studies, \cite{b1} employed a Hidden Markov model for NER in the Hindi Language, utilizing 12 tags and achieving 97.14\% accuracy. \cite{b2} proposed novel CNN and LSTM architectures for NER, working with five English and one Chinese datasets. Additionally, researchers have demonstrated that combining deep learning with active learning can enhance the performance of NLP models. Another notable work by \cite{b3} introduced the first open neural NLP model for German NER tasks.

Evaluation of state-of-the-art machine learning approaches for named entity recognition on the Semantic Web was conducted by \cite{b4}. The Multilayer Perceptron performed best in terms of f-score, with 0.04\% higher recall than Random Forest. However, poorer results were observed in the entity-based evaluation, where MLP ranked second to Functional Trees.

Two novel neural architectures, namely LSTMs, and StackLSTM, have been developed to improve NER performance across multiple languages, including English. The LSTM-CRF model outperformed other systems, even those utilizing externally labeled data such as gazetteers. Similarly, the StackLSTM model demonstrated superior performance compared to previous models that lacked external features, as reported in \cite{b6}.

Challenges in performing NER in Indian languages using a Hidden Markov Model (HMM) were discussed by \cite{b5}. The researchers handled a total of seven named entity tags and achieved accuracies of 86\%  for Hindi, 76\% for Marathi, and 65\% for Urdu. They experimented with Conditional Random Fields and Maximum Entropy models, altering the feature set to identify the most effective feature set for each model \cite{b7}.

In \cite{b8}, the authors investigated three neural network-based models for Indonesian Named Entity Recognition. The BiLSTM-CNNs + pre-trained word2vec embedding model exhibited strong performance with an f1 score of 71.37

\cite{b9} focused on NER in Twitter, constructing a new dataset called Tweet-NER7, which contained annotated entities of seven types across 11,382 tweets. In the analysis, three crucial temporal factors were taken into account: the deterioration of NER models in the short term, approaches for adapting a language model across varying timeframes, and the potential use of self-labeling as a substitute in scenarios where recently labeled data is scarce.

Lastly, \cite{b10} introduced L3Cube-MahaNER, a significant gold Marathi NER dataset with 8 target labels. The dataset was benchmarked on various CNN, LSTM, and Transformer-based models such as MahaBERT, IndicBERT, mBERT, and XLM-RoBERTa. Similarly, in \cite{litake2023mono}, the authors focus on NER in low-resource Indian languages like Marathi and Hindi. The study investigates different transformer-based models, such as base-BERT, AlBERT, and RoBERTa, and assesses their effectiveness on publicly accessible NER datasets in Hindi and Marathi. The data reveal that the MahaRoBERTa model obtains greater performance in Marathi NER, whereas the XLM-RoBERTa model outperforms others in Hindi NER.
 
\section{Dataset Details}\label{sec2}

\subsection{Dataset Introduction}\label{subsec2}
The Dataset used in this work is the L3Cube-MahaNER \cite{b10}, which is the first major gold standard Marathi NER corpus. Consisting of 25,000 sentences, the dataset is in Marathi Language. L3Cube-MahaCorpus \cite{b17} is used as the base for these sentences which are monolingual and extracted from news domains. 

The dataset has predefined train, test, and validation splits. The training dataset consists of 21500 sentence counts along with a 26502 tag count, the test dataset consists of 2000 sentences and a 2424 tag count and the validation dataset consists of 1500 sentences and 1800 tag count. 

\begin{figure}[htbp]
\centerline{\includegraphics[scale=0.8]{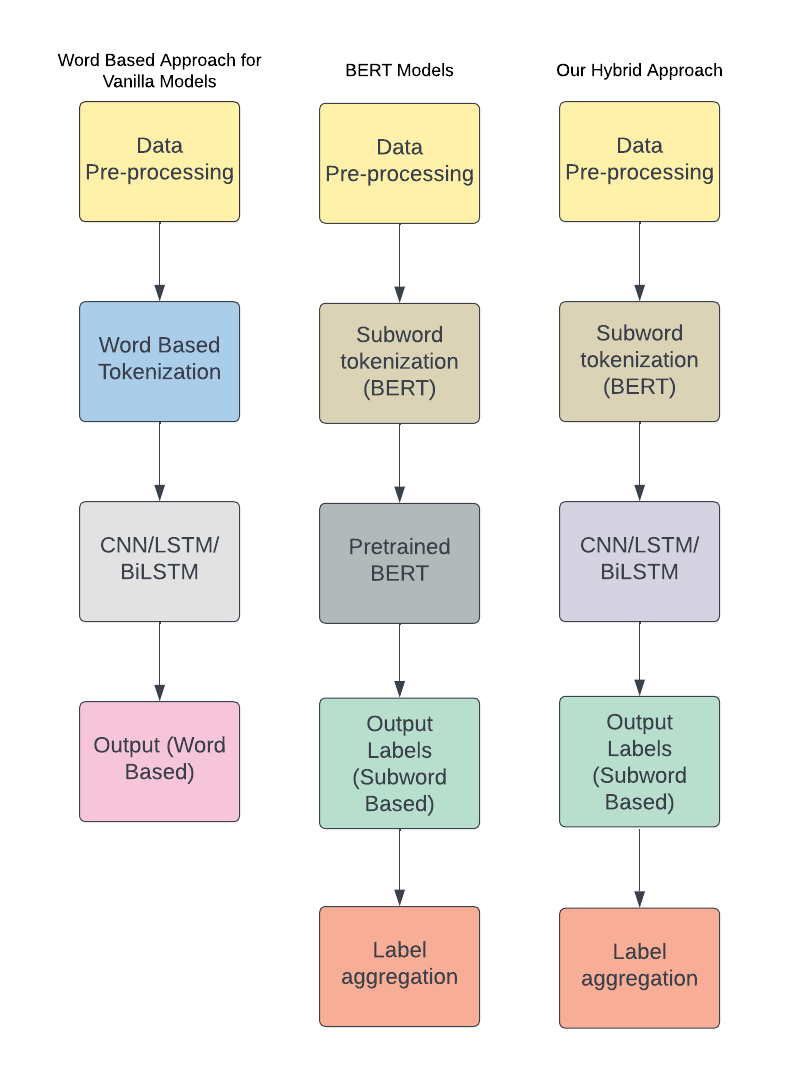}}
\caption{There are three approaches: The first column consists of Word Based Approach for Vanilla Models \cite{b10}, the second column consists of BERT Models \cite{b10} and the third column consist of our hybrid approach which additionally includes the allocation of labels to the subtokens before training and clubbing of labels generated by the models for the subtokens to the root tokens.}
\label{fig:proposed}
\end{figure}

\section{Proposed Methodology}\label{sec3}

\subsection{Models Architectures}\label{subsec2}
Deep learning has revolutionized NLP and become the go-to approach for many tasks. Deep learning models are capable of automatically learning complicated patterns and representations from massive volumes of data, making them particularly successful for applications such as named entity recognition. Depending on the task and the type of data being used, deep learning models can have a broad range of architectures.
\newline
\newline
\textbf{CNN:}
The utilization of Convolutional Neural Network (CNN) models for named entity recognition (NER) is advantageous due to their ability to extract local context, handle word order invariance, perform feature composition, enable effective parameter sharing, and deliver strong performance on local context tasks.

In CNN-based NER models, the input text is typically converted into a sequence of word embeddings, representing each word in a high-dimensional space. In this particular model, a single 1D convolutional layer is employed, with the word embeddings having a dimension of 300. These embeddings are typically trained as part of the NER model training procedure. The 1D convolutional layer, which utilizes the 'relu' activation function and has 512 filters with a kernel size of 3, receives the embeddings after that. The output of the Conv1D layer is then fed into a dense layer that has the same size as the output layer and uses the 'softmax' activation function. As there are 8 classes, the model generates 8 output labels. The 'rmsprop' optimizer is employed in the training process.
\newline
\newline
\textbf{LSTM:}
The utilization of Long Short-Term Memory (LSTM) models for named entity recognition (NER) is advantageous due to their ability to process sequential information, comprehend context, handle variable-length sequences, and address the issue of vanishing gradients.

Similar to the CNN model, the LSTM model also follows a similar architecture. In this case, the model comprises a single LSTM layer with word embeddings of 300 dimensions. Following this layer, a dense layer with properties resembling those of the CNN model is present and contains 512 filters.
\newline
\newline
\textbf{BiLSTM:}
The usage of Bidirectional Long Short-Term Memory (BiLSTM) models for named entity recognition (NER) is advantageous due to their ability to provide contextual understanding, improve representation learning, handle ambiguity robustly, fuse features, and handle out-of-order entities.

BiLSTM's model architecture is comparable to that of CNN's, with a BiLSTM layer in place of the single 1D convolutional layer. This particular model uses a 512-hidden-unit BiLSTM layer and an embedding vector with 300 dimensions. There are 16 batches total.

While the CNN, LSTM, and BiLSTM architectures are typically considered pre-existing options, our research distinguishes itself primarily through the approach taken in the tokenization operation.

\subsection{Tokenization}\label{subsec2}
Tokenization involves breaking sentences or paragraphs into tokens, which are then used for model training. Different types of tokenization exist, such as word-based, subword-based, and character tokenization. While word tokenizers were traditionally used with CNN/LSTM models, our research employs subword-based tokenizers in conjunction with these models. The proposed flow is shown in Figure \ref{fig:proposed}.

\textbf{Subword based Tokenizers:}
Subword tokenization enhances NER by effectively handling unfamiliar words, capturing morphological variations, improving generalization, and enabling cross-lingual transfer learning. Breaking words into smaller sub-tokens enables the model to identify and infer from hidden word components. This detailed approach is particularly beneficial for entities with morphological alterations, as it captures prefixes, suffixes, and stem variations. Moreover, it enables generalization across similar terms and facilitates knowledge transfer across languages. Overall, subword tokenization broadens the scope of NER, accommodates morphological variations, and enhances performance specifically in the Marathi language. The core methodology described in this work is to integrate subword tokenizers with vanilla CNN/LSTM models. We specifically focus on BERT-based subword tokenizers.

\begin{figure}[htbp]
\centerline{\includegraphics[scale=0.5]{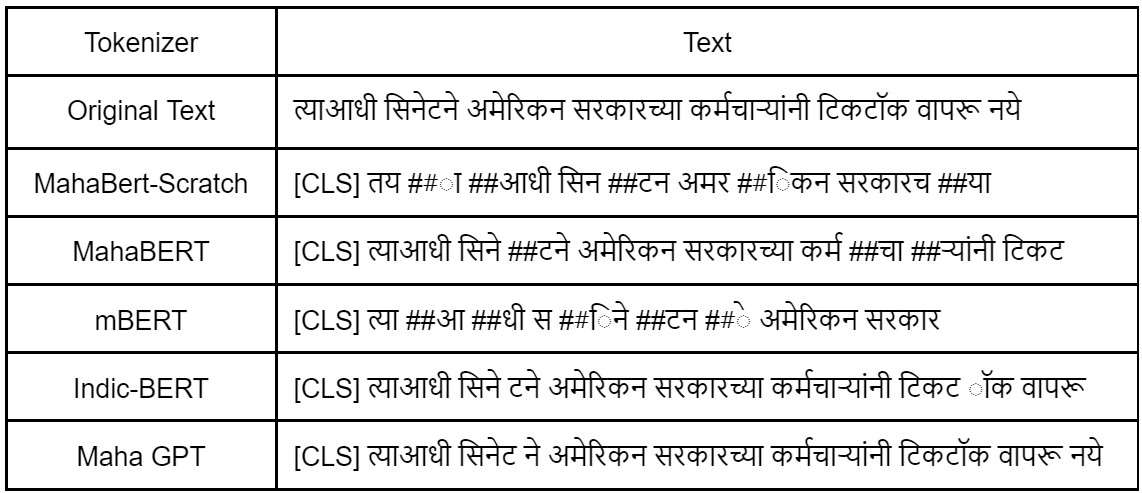}}
\caption{Tokens generated by various tokenizers.}
\label{fig:tokens}
\end{figure}

The different types of Marathi subword-based tokenizers based on BERT are as follows:
\begin{itemize}
    \item MahaBERT-Scratch Tokenizer\footnote{\url{https://huggingface.co/l3cube-pune/marathi-bert-scratch}}  \cite{b17,b16}
    \item MahaBERT Tokenizer\footnote{\url{https://huggingface.co/l3cube-pune/marathi-bert-v2}} \cite{b17,b16}
    \item mBERT Tokenizer\footnote{\url{https://huggingface.co/bert-base-multilingual-cased}} \cite{devlin-etal-2019-bert}
    \item IndicBERT Tokenizer\footnote{\url{https://huggingface.co/ai4bharat/indic-bert}} \cite{kakwani2020indicnlpsuite}
    \item Marathi-GPT Tokenizer\footnote{\url{https://huggingface.co/l3cube-pune/marathi-gpt}} \cite{b17}
\end{itemize}

Eventually, these BERT-based tokenizers were used separately for each model and the metrics were calculated. A sample sentence split for each model is shown in Figure \ref{fig:tokens}.

There were two major requirements encountered during the sub-tokenization before training and after testing the models: -
\newline
\textbf{Requirement before training the model:} The root word may be split into multiple sub-word tokens. So the entity label of the root token needs to be passed to its corresponding sub-word tokens.
\newline
\textbf{Requirement after testing the model:} The labels generated by the model for the multiple sub-tokens need to be passed to the single root token.

\begin{table}
\begin{center}
\begin{adjustbox}{width=350pt,center}
    \begin{tabular}{|c|c|c|c|c|c|c|c|c|c|c|c|c|}
    
    \hline
    \multicolumn{1}{|c|}{  } &
    \multicolumn{3}{|c|}{F1-Score} &
    \multicolumn{3}{|c|}{Precision} &
    \multicolumn{3}{|c|}{Recall} &
    \multicolumn{3}{|c|}{Accuracy} \\
    \hline
    
    Tokenizer/Model & CNN & LSTM & BiLSTM & CNN & LSTM & BiLSTM & CNN & LSTM & BiLSTM & CNN & LSTM & BiLSTM \\
    \hline
    
    Word-Based & 79.5 & 74.9 & 80.4 & 82.1 & \textbf{84.1} & 83.3 & 77.4 & 68.5 & 77.6 & \textbf{97.28} & 94.89 & 94.99 \\
    
    MahaBert-Scratch & 76.4 & 65.2 & 75.2 & 81.3 & 76.9 & 83.9 & 72.6 & 58.8 & 69.9 & 95.0 & 93.0 & 95.0 \\
    
    \textbf{MahaBERT} & \textbf{82.1} & \textbf{76.0} & \textbf{82.0} & \textbf{84.9} & 79.9 & 83.9 & \textbf{79.9} & \textbf{73.0} & 80.6 & 96.0 & \textbf{95.0} & \textbf{96.0} \\
    
    mBert & 66.4 & 50.0 & 75.4 & 78.1 & 72.3 & 78.0 & 59.4 & 41.7 & \textbf{73.2} & 93.0 & 91.0 & 94.0 \\
    
    Indic-Bert & 81.8 & 75.6 & 81.2 & 83.8 & 82.1 & 80.4 & 79.9 & 70.8 & 82.7 & 96.0 & 95.0 & 95.0 \\
    
    Marathi GPT & 80.4 & 58.2 & 74.5 & 81.4 & 80.9 & \textbf{84.0} & 79.6 & 56.4 & 68.9 & 95.0 & 94.0 & 95.0 \\
    \hline
\end{tabular}
\end{adjustbox}
\vspace{0.1cm}
\caption{\label{result-tab} Comparison table of F1 score, Precision, Recall, and Accuracy of word-based tokenizer and subword-based tokenizers. Note that model name in the first columns represents the tokenizer type}
\end{center}
\end{table}

\begin{table}
\begin{center}
    \begin{tabular}{|c|c|}
        \hline
        Tokenizer/Model & F1-Score \\
        \hline
        CNN + word tokenizer & 79.5\\
        MahaBert & 86.8\\
        CNN + MahaBert tokenizer & 82.1\\
        \hline
    \end{tabular}
\vspace{0.1cm}
\caption{\label{base-result-tab} F1 scores of CNN model with word-based tokenizer, MahaBERT model, and CNN model with MahaBERT tokenizer.}
\end{center}
\end{table}

\begin{figure}[htbp]
\centerline{\includegraphics[scale=0.5]{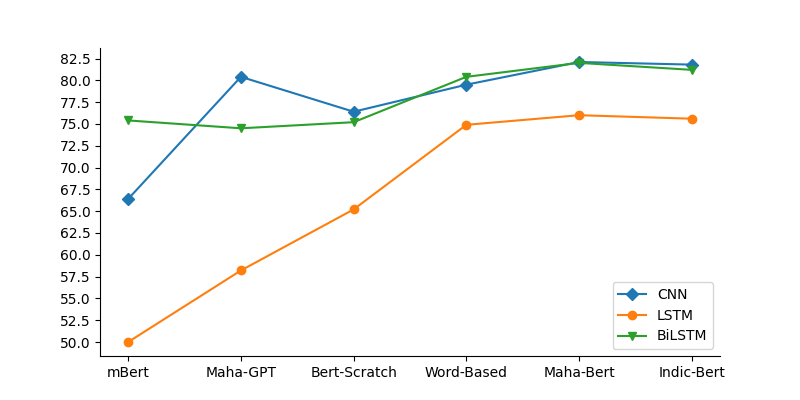}}
\caption{F1 score for all sub-tokenizers}
\label{fig:tokenizers}
\end{figure}

\section{Results}\label{sec4}

In this research, we conducted experiments on Marathi, a low-resource language, using shallow vanilla models such as CNN, LSTM, and BiLSTM. Our goal was to conduct named entity recognition on the MahaNER dataset. To bridge the gap between deep pre-trained models like BERT transformers and non-pre-trained shallow deep learning models, we augmented these low-level models with subword integration. We employed BERT-based tokenizers including MahaBERT-Scratch, MahaBERT, mBERT, MuRIL, and indicBERT tokenizers. In this section, we present the achieved recall, precision, F1 scores, and accuracy (in percentage) after training the models. We also provide the results obtained using word-based tokenizers from previous research conducted on the same dataset. The performance of different tokenizer types is shown in Table \ref{result-tab} and Figure \ref{fig:tokenizers}. The baseline comparison with pre-trained MahaBERT model and word-based vanilla CNN model is shown in Table \ref{base-result-tab}.

Based on the evaluation, we observe that the MahaBERT tokenizer produces the highest F1 scores among all benchmarked sub-tokenizers and the word-based tokenizer: 82.1\% for CNN, 76.0\% for LSTM, and 82\% for BiLSTM. We also consider the F1 scores of the word-based tokenizer as the baseline, 79.5\% for CNN, 74.9\% for LSTM, and 80.4\% for BiLSTM. Thus the sub tokenizers contribute to a significant improvement in the F1-score of the single-layer vanilla models. Subword tokenization expands the application of NER, accommodates morphological variations, and enhances performance in the Marathi language. As a result, our hybrid approach distinguishes itself from the traditional word-based approach and achieves results comparable to those obtained with BERT models.

\section{Limitations}\label{sec4}
\begin{itemize}
    \item The application of subword tokenization increases the length of the input sequence, resulting in longer training times.
    \item Although we have reduced the difference between vanilla models and BERT based models to half, the scope for further improvements are still possible.
\end{itemize}

\section{Conclusion}\label{sec4}
In this work, our focus was on performing named entity recognition (NER) in Marathi, a low-resource language. We addressed the challenge of bridging the gap between high-level models like BERT transformers and low-level models such as CNN, LSTM, and BiLSTM. We propose a hybrid approach to integrate BERT based sub-word tokenizers into the vanilla CNN and LSTM models. We compare different tokenizers from monolingual models like MahaBERT, MahaGPT, MahaBERT-Scratch and multilingual models like IndicBERT, mBERT.
Among the different models, the CNN model achieved the highest F1-score using the MahaBERT tokenizer. The increased accuracy and effectiveness of these low-level models can lead to reduced computation costs for NLP applications where NER serves as a fundamental operation.

\section*{Acknowledgements}\label{sec4}
We would like to express our sincere gratitude towards the L3Cube mentorship program and our mentor for their continual support and guidance. We are grateful to Pune Institute of Computer Technology for encouraging and supporting us throughout the research period. The issue statement and ideas provided in this work are from L3Cube and its mentors and are a part of the L3Cube-MahaNLP project\cite{joshi2022l3cube_mahanlp}.

\bibliographystyle{splncs03}
\bibliography{main}

\begin{thebibliography}{10}
\providecommand{\url}[1]{\texttt{#1}}
\providecommand{\urlprefix}{URL }

\bibitem{b1}
Chopra, D., Joshi, N., Mathur, I.: Named entity recognition in hindi using
  hidden markov model. 2016 Second International Conference on Computational
  Intelligence \& Communication Technology (CICT) pp. 581--586 (2016)

\bibitem{devlin-etal-2019-bert}
Devlin, J., Chang, M.W., Lee, K., Toutanova, K.: {BERT}: Pre-training of deep
  bidirectional transformers for language understanding. In: Proceedings of the
  2019 Conference of the North American Chapter of the Association for
  Computational Linguistics: Human Language Technologies, Volume 1 (Long and
  Short Papers). pp. 4171--4186. Association for Computational Linguistics,
  Minneapolis, Minnesota (Jun 2019), \url{https://aclanthology.org/N19-1423}

\bibitem{b3}
Frei, J., Kramer, F.: {GERNERMED} - an open german medical {NER} model. CoRR
  abs/2109.12104 (2021), \url{https://arxiv.org/abs/2109.12104}

\bibitem{10.1007/978-981-16-3690-5_73}
Joshi, R., Joshi, R.: Evaluating input representation for language
  identification in hindi-english code mixed text. In: ICDSMLA 2020. pp.
  795--802. Springer Singapore, Singapore (2022)

\bibitem{b16}
Joshi, R.: L3cube-hindbert and devbert: Pre-trained bert transformer models for
  devanagari based hindi and marathi languages. arXiv preprint arXiv:2211.11418
   (2022)

\bibitem{b17}
Joshi, R.: L3cube-mahacorpus and mahabert: Marathi monolingual corpus, marathi
  bert language models, and resources. In: Proceedings of the WILDRE-6 Workshop
  within the 13th Language Resources and Evaluation Conference. pp. 97--101
  (2022)

\bibitem{joshi2022l3cube_mahanlp}
Joshi, R.: L3cube-mahanlp: Marathi natural language processing datasets,
  models, and library. arXiv preprint arXiv:2205.14728  (2022)

\bibitem{kakwani2020indicnlpsuite}
Kakwani, D., Kunchukuttan, A., Golla, S., Gokul, N., Bhattacharyya, A., Khapra,
  M.M., Kumar, P.: Indicnlpsuite: Monolingual corpora, evaluation benchmarks
  and pre-trained multilingual language models for indian languages. In:
  Findings of the Association for Computational Linguistics: EMNLP 2020. pp.
  4948--4961 (2020)

\bibitem{kudo2018subword}
Kudo, T.: Subword regularization: Improving neural network translation models
  with multiple subword candidates. In: Proceedings of the 56th Annual Meeting
  of the Association for Computational Linguistics (Volume 1: Long Papers). pp.
  66--75 (2018)

\bibitem{kudo2018sentencepiece}
Kudo, T., Richardson, J.: Sentencepiece: A simple and language independent
  subword tokenizer and detokenizer for neural text processing. arXiv preprint
  arXiv:1808.06226  (2018)

\bibitem{b6}
Lample, G., Ballesteros, M., Subramanian, S., Kawakami, K., Dyer, C.: Neural
  architectures for named entity recognition. arXiv preprint arXiv:1603.01360
  (2016)

\bibitem{b2}
Li, J., Sun, A., Han, J., Li, C.: A survey on deep learning for named entity
  recognition. IEEE Transactions on Knowledge and Data Engineering  34(1),
  50--70 (2020)

\bibitem{litake2023mono}
Litake, O., Sabane, M., Patil, P., Ranade, A., Joshi, R.: Mono versus
  multilingual bert: A case study in hindi and marathi named entity
  recognition. In: Proceedings of 3rd International Conference on Recent Trends
  in Machine Learning, IoT, Smart Cities and Applications: ICMISC 2022. pp.
  607--618. Springer (2023)

\bibitem{b10}
Litake, O., Sabane, M.R., Patil, P.S., Ranade, A.A., Joshi, R.: L3cube-mahaner:
  A marathi named entity recognition dataset and bert models. In: Proceedings
  of the WILDRE-6 Workshop within the 13th Language Resources and Evaluation
  Conference. pp. 29--34 (2022)

\bibitem{b7}
Manamini, S., Ahamed, A., Rajapakshe, R., Reemal, G., Jayasena, S., Dias, G.,
  Ranathunga, S.: Ananya-a named-entity-recognition (ner) system for sinhala
  language. In: 2016 Moratuwa Engineering Research Conference (MERCon). pp.
  30--35. IEEE (2016)

\bibitem{b11}
Patil, P., Ranade, A., Sabane, M., Litake, O., Joshi, R.: L3cube-mahaner: A
  marathi named entity recognition dataset and bert models. arXiv preprint
  arXiv:2204.06029  (2022)

\bibitem{sennrich2016neural}
Sennrich, R., Haddow, B., Birch, A.: Neural machine translation of rare words
  with subword units. In: Proceedings of the 54th Annual Meeting of the
  Association for Computational Linguistics (Volume 1: Long Papers). pp.
  1715--1725 (2016)

\bibitem{DBLP:journals/corr/ShenYLKA17}
Shen, Y., Yun, H., Lipton, Z.C., Kronrod, Y., Anandkumar, A.: Deep active
  learning for named entity recognition. CoRR  abs/1707.05928 (2017),
  \url{http://arxiv.org/abs/1707.05928}

\bibitem{b5}
Singh, J., Joshi, N., Mathur, I.: Part of speech tagging of marathi text using
  trigram method. arXiv preprint arXiv:1307.4299  (2013)

\bibitem{b4}
Speck, R., Ngonga~Ngomo, A.C.: Ensemble learning for named entity recognition.
  In: Mika, P., Tudorache, T., Bernstein, A., Welty, C., Knoblock, C.,
  Vrande{\v{c}}i{\'{c}}, D., Groth, P., Noy, N., Janowicz, K., Goble, C. (eds.)
  The Semantic Web -- ISWC 2014. pp. 519--534. Springer International
  Publishing, Cham (2014)

\bibitem{b8}
Sukardi, S., Susanty, M., Irawan, A., Putra, R.F.: Low complexity named-entity
  recognition for indonesian language using bilstm-cnns. In: 2020 3rd
  International Conference on Information and Communications Technology
  (ICOIACT). pp. 137--142. IEEE (2020)

\bibitem{b9}
Ushio, A., Neves, L., Silva, V., Barbieri, F., Camacho-Collados, J.: Named
  entity recognition in twitter: A dataset and analysis on short-term temporal
  shifts. arXiv preprint arXiv:2210.03797  (2022)

\end{thebibliography}

\end{document}